\documentclass{article}
\usepackage{amsmath}

\usepackage{PRIMEarxiv}

\usepackage[utf8]{inputenc} 
\usepackage[T1]{fontenc}    
\usepackage{hyperref}       
\usepackage{url}            
\usepackage{booktabs}       
\usepackage{amsfonts}       
\usepackage{nicefrac}       
\usepackage{microtype}      
\usepackage{lipsum}
\usepackage{graphicx}
\usepackage{lineno}
\usepackage[utf8]{inputenc} 
\usepackage[T1]{fontenc}    
\usepackage{hyperref}       
\usepackage{url}            
\usepackage{booktabs}       
\usepackage{amsfonts}       
\usepackage{nicefrac}       
\usepackage{microtype}      
\usepackage{xcolor}         

\usepackage{amssymb}
\usepackage{times}
\usepackage{latexsym}
\usepackage{float}
\usepackage{multirow}
\usepackage{amsmath}
\usepackage{inconsolata}
\usepackage{graphicx}
\usepackage{subcaption}
\usepackage{algorithm}
\usepackage{algorithmic}
\usepackage{wrapfig}
\graphicspath{{media/}}     

\title{One Head, Many Models: Cross-Attention Routing for Cost-Aware LLM Selection}

\author{
Roshini Pulishetty$^{1,2}$\thanks{These authors contribute equally.} \quad Mani Kishan Ghantasala$^{1}$\footnotemark[1] \quad Keerthy Kaushik Dasoju$^{1}$\footnotemark[1] \quad Niti Mangwani$^{1}$\footnotemark[1] \quad \\
\textbf{Vishal Garimella}$^{1}$ \quad \textbf{Aditya Mate}$^{2}$ \quad \textbf{Somya Chatterjee}$^{2}$ \quad \textbf{Yue Kang}$^{2}$ \\ \textbf{Ehi Nosakhare}$^{2}$ \quad \textbf{Sadid Hasan}$^{2}$ \quad \textbf{Soundar Srinivasan}$^{2}$ \\
$^1$University of Massachusetts, Amherst \quad $^2$Microsoft\\
\texttt{\{rpulishetty,mghantasala,kdasoju,nmangwani,vishalg\}@umass.edu}\\
\texttt{\{adityamate,sharmasomya,yuekang,ehnosakh,sadidhasan,sosrini\}@microsoft.com}
}

\begin{document}
\maketitle

\begin{abstract}
The proliferation of large language models (LLMs) with varying computational costs and performance profiles presents a critical challenge for scalable, cost-effective deployment in real-world applications. We introduce a unified routing framework that leverages a single-head cross-attention mechanism to jointly model query and model embeddings, enabling dynamic selection of the optimal LLM for each input query. Our approach is evaluated on RouterBench, a large-scale, publicly available benchmark encompassing diverse LLM pools and domains. By explicitly capturing fine-grained query-model interactions, our router predicts both response quality and generation cost, achieving up to 6.6\% improvement in Average Improvement in Quality (AIQ) and 2.9\% in maximum performance over existing routers. To robustly balance performance and cost, we propose an exponential reward function that enhances stability across user preferences. The resulting architecture is lightweight, generalizes effectively across domains, and demonstrates improved efficiency compared to prior methods, establishing a new standard for cost-aware LLM routing.
\end{abstract}


\section{Introduction}

\begin{wrapfigure}{r}{0.6\textwidth}
    \centering
    \includegraphics[width=\linewidth]{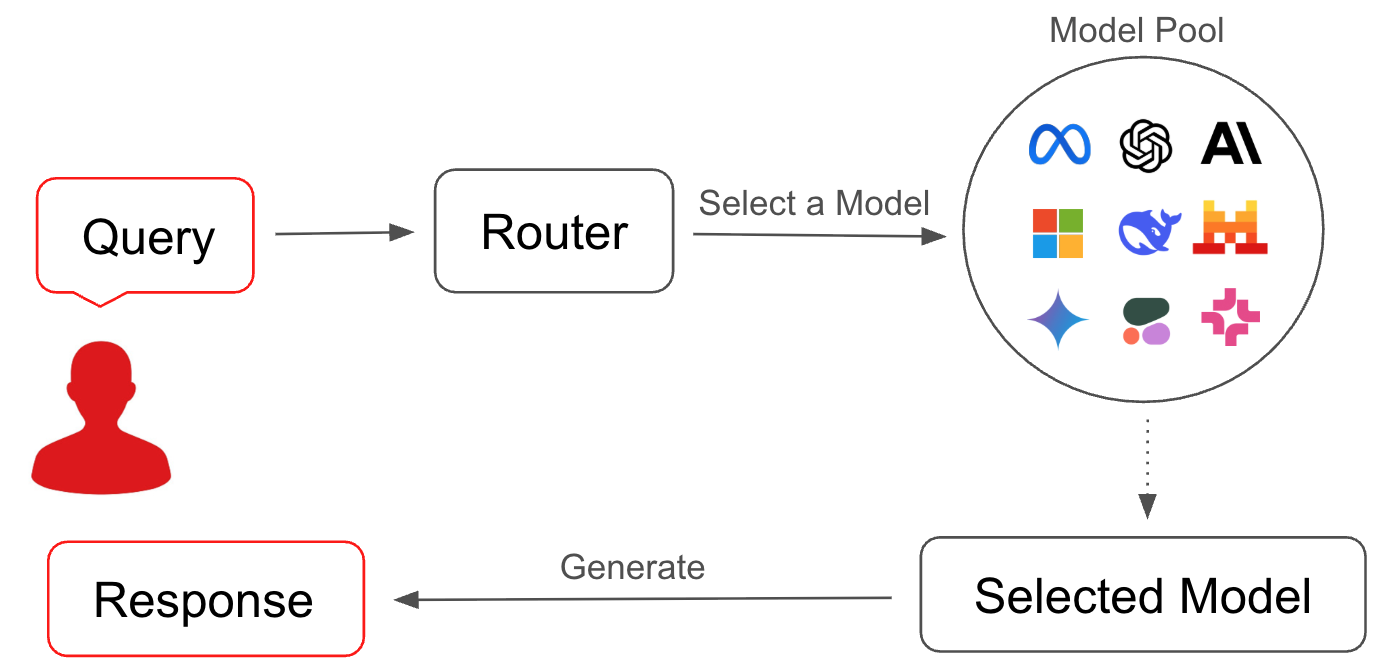}
    \caption{LLM Router selects an appropriate model for each query to route to}
    \label{fig:router}
\end{wrapfigure}
The rise of large language models (LLMs) has advanced reasoning, summarization, and code generation. Yet, the wide range of options, from lightweight, cost-efficient models (e.g., Mistral 7B~\cite{jiang2023mistral}) to powerful but costly ones (e.g., GPT-4~\cite{achiam2023gpt}), creates a core challenge: selecting the right model per query to balance response quality and cost. This challenge is critical for hyperscalers, where both efficiency and user experience are paramount.


Recent work has proposed diverse strategies for LLM routing to balance quality and cost. Classification-based methods predict the best model from static query features~\cite{ong2024routellm,liu2024optllm,somerstep2025carrot}, but assume fixed model sets and rely on pre-computed metrics, limiting adaptability. Reinforcement learning–based approaches learn dynamic policies~\cite{li2025llmbandit,sikeridis2024pickllm}, yet require many interactions to converge and suffer in cold-start scenarios. Training-free and heuristic methods such as LLM-BLENDER\cite{jiang2024llmblender} boost accuracy by combining outputs, but incur the overhead of querying all models. Similarity-based approaches~\cite{stripelis2024tensoropera,pichlmeier2024domainaware} leverage embeddings or domain classifiers, but often depend on static experts or domain-specific assumptions.

A common limitation of prior approaches is treating queries and models as independent, with routing based on query-only features or model-agnostic heuristics. Our method instead models query–model interactions via cross-attention, allowing the router to assess how a model will perform on a given query, enabling flexible, domain-agnostic routing.

Empirical results on RouterBench demonstrate that our attention-based router consistently achieves higher AIQ score across different LLM pools, outperforming traditional baselines- KNN, MLP and SVM routers by an average of $23.85\%$, $3.34\%$ and $27.33\%$ respectively. Ablation studies confirm that attention-based predictors consistently outperform regression and MLP variants, with up to 10.9\% higher AIQ and 9.7\% higher maximum performance. These findings highlight the importance of modeling query--model interactions for scalable and efficient LLM\footnote{As the definition of ‘large’ language models evolves, we use ‘LLM’ to refer to the models included in our experimental pool.} routing.



\section{Related Work}\label{sec:relatedwork}

As large language models (LLMs) proliferate across domains and deployment settings, the challenge of selecting the most appropriate model for a given query has become central to efficient and effective LLM usage. Prior work on LLM routing has largely focused on optimizing for cost, quality, or adaptability — but often treats these objectives in isolation or relies on rigid assumptions about model behavior. Our work builds on this foundation by proposing a unified, interaction-based routing framework that jointly models query and LLM characteristics to make informed, flexible routing decisions.

Early approaches to LLM routing framed the problem as a classification task, where a supervised model predicts the best LLM for a query. These methods, such as RouteLLM \cite{ong2024routellm} and OptLLM \cite{liu2024optllm}, demonstrated that static query features could be used to reduce reliance on expensive models.  Recent work, CARROT \cite{somerstep2025carrot}, predicts both cost and accuracy to select the most cost-effective model, and deduces minimax optimality under certain assumptions. However, they often assume a fixed set of models and rely on pre-computed performance metrics, limiting their ability to generalize to new domains or adapt to evolving model pools.

In parallel, reinforcement learning-based methods introduced dynamic routing policies that adapt over time. Methods like LLM Bandit \cite{li2025llmbandit} and PickLLM \cite{sikeridis2024pickllm} use online feedback to refine model selection strategies. While these methods offer adaptability, they typically require many interactions to converge and struggle with cold-start scenarios — a critical limitation in real-world deployments where immediate routing decision is essential. Causal LLM Routing \cite{tsiourvas2024causalllm} learns routing policies from observational data via end-to-end regret minimization, avoiding costly full-feedback datasets, but still depends on rich historical logs and omits explicit query–model interaction modeling.

Immediate routing decisions can be achieved by exploring training-free or heuristic-based routing, aimed to reduce overhead by avoiding model training altogether. Eagle \cite{zhao2024eagle} and Universal Model Routing \cite{jitkrittum2025universal} use ranking systems or unsupervised clustering to guide routing decisions. These methods are appealing for their simplicity and low cost, but often lack the granularity needed to capture nuanced differences in model behavior, especially for complex or ambiguous queries. From another angle, LLM-BLENDER \cite{jiang2024llmblender} ensembles multiple LLMs by ranking and fusing their generated outputs using a pairwise ranking module and a generative fusion module. While effective, this post-generation approach requires outputs from all candidate models, in contrast to our pre-generation routing perspective, which selects a single model before inference.


Across these diverse approaches, a common limitation emerges: most methods treat the query and model as independent entities, relying on either query-only features or model-agnostic heuristics. In contrast, our approach explicitly models the interaction between query and model embeddings, allowing the router to reason about how a specific query might perform on a specific model. By learning to predict both quality and cost in a unified framework, our method supports flexible, domain-agnostic routing that adapts to new models and tasks with minimal supervision. This problem was also studied by~\cite{panda2025adaptive}, a concurrent and parallel work that appeared online only a few days apart from ours, and our effort is independent in nature.

\section{Method} \label{sec:method}

\paragraph{Problem Formulation}
Given a pool of LLMs $\mathcal{M} = \{M_1, M_2,..M_K\}$ and user query space $\mathcal{Q}$, our goal is to design an LLM router (Figure \ref{fig:router}) as a decision-making agent $\Pi:\mathcal{Q}\rightarrow\mathcal{M}$, mapping queries to models under response uncertainty. The router is designed to balance the trade-off between performance and cost, optimizing the competing objectives of maximizing response quality while minimizing resource usage. Our guiding principle is that a query should only be routed to an expensive model if all cheaper models fail to give a promising response and the user is willing to pay the additional cost.


\paragraph{Predictor-based Routing Framework}
In this work, we employ a predictor-based LLM routing framework and propose attention as an effective architecture to estimate the response quality or the generation cost of candidate models. Based on these estimates, the framework selects the most suitable model by incorporating user's willingness to pay in the reward function, thus decoupling predictors training from user's parameter. To ensure scalability across the model pool, we design a dual-predictor framework where one predictor estimates performance of all models, while the other estimates generation cost. Intuitively, as the user’s willingness to pay increases, the framework places greater emphasis on response quality while discounting the cost factor. Later, we perform a systematic study on choosing an appropriate reward function for this framework \ref{sec:results-rewards}.


\begin{wrapfigure}{r}{0.6\textwidth}
    \centering
    \includegraphics[width=\linewidth]{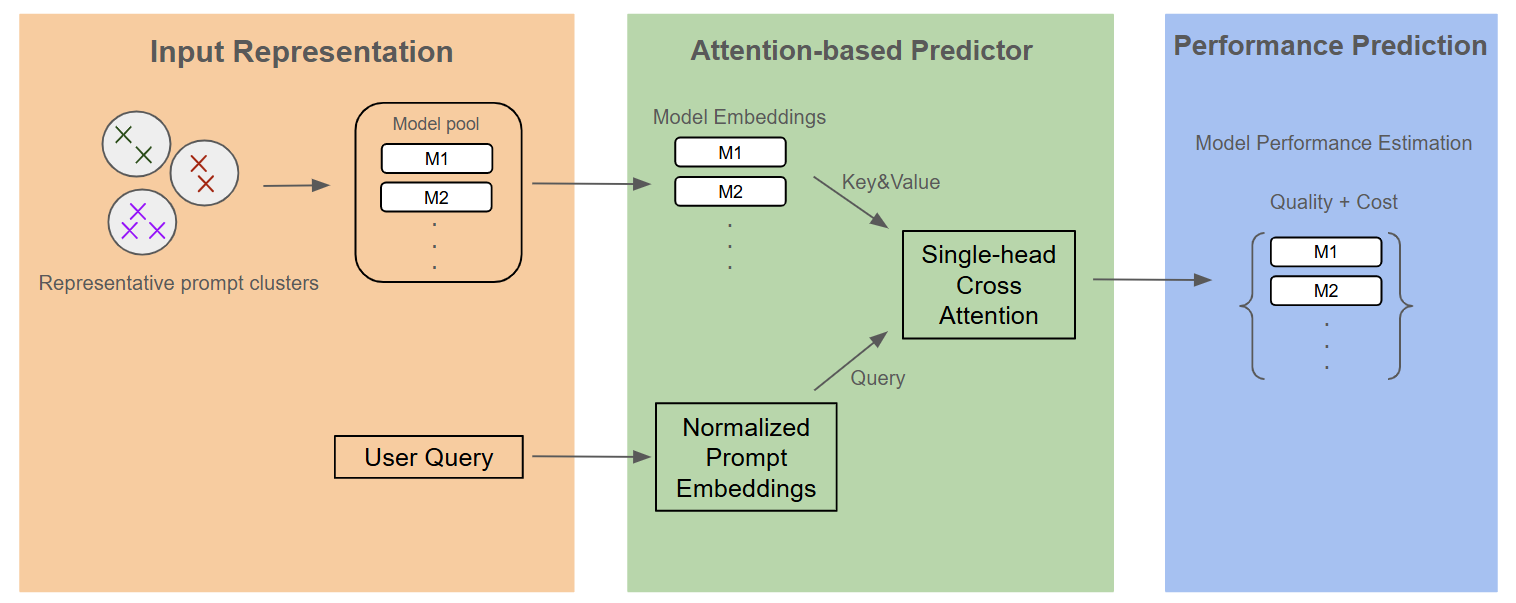}
    \caption{Single-head cross-attention block}
    \label{fig:mha_attention}
\end{wrapfigure}

\paragraph{Attention-Based Predictors.} We propose similarity-based routing using a single-head cross-attention which encodes the incoming prompts as queries and LLM representations (See Appendix \ref{para:llm-representations}) as keys and values. We presume query ($\Vec{\mathbf{q}}$) captures level of prompt's complexity in multi-dimensions, while key ($\Vec{\mathbf{k}}$) and value ($\Vec{\mathbf{v}}$) express LLM's expertise in these dimensions. 
This predictor captures query–model interactions through attention, enabling it to estimate expected performance and generation cost of response from each model for a given prompt. 
\[
\text{Attention}(\Vec{\mathbf{q}}, \Vec{\mathbf{k}}, \Vec{\mathbf{v}}) = \text{softmax}\left( \frac{\Vec{\mathbf{q}}. \Vec{\mathbf{k}}^T}{\sqrt{d_v}} \right) \Vec{\mathbf{v}}.
\]

This similarity-based routing framework offers several advantages. By decoupling model representation creation from the training process, it remains adaptive to an evolving model pool with minimal retraining effort. The cross-attention module maps queries and model embeddings into a shared latent space, without restraining sizes of prompt and LLM representations. As a second-order similarity mechanism, attention could potentially capture richer and more nuanced interactions between queries and models than dot product or cosine similarity. Further, its inherent parallelism makes the router scalable to heavy query traffic, effectively handle larger LLM pools and batched queries.

To establish other predictor variants, we also explore regression (Reg) and neural network-based (FCN) predictors.

\textbf{Predictor Variants.} We primarily explored $2$ architectural variants: 
\begin{enumerate}
    \item \textbf{Regression-Based Predictors:} A linear regressor learns a best-fit mapping from query embeddings to model performance and cost. While interpretable and computationally efficient, this approach is limited to capturing linear relationships. $$QX_1=S\in\mathcal{R}^{n\times q} \ \ \ QX_2=C\in\mathcal{R}^{n\times q}$$
    \item \textbf{Neural Network Predictors:} Fully connected networks (two-layer and three-layer MLPs) map query embeddings to performance and cost predictions for all candidate models, i.e., $f(q;\theta)=S\in\mathbb{R}^m$ and $g(q;\theta)=C\in\mathbb{R}^m$. Unlike regression-based predictors, parameters are shared across predictions for different models.
\end{enumerate}

\paragraph{Model Representations Augmentation}
We further augment predictors input by concatenating model embeddings to the query embeddings and feed it to different predictor architectures, denoted as Reg-emb, 2FCN-emb, and 3FCN-emb in Appendix~\ref{sec:predictor-routing-extra-results}. Decoupling model representation construction from predictor training enables dynamic addition and removal of models at the inference time. Given a query embedding and a model embedding, we concatenate the two vectors to form the predictor’s input. The predictor then outputs the predicted performance or cost of the model on that query.

\section{Evaluation Methodology}

\paragraph{Data.} We evaluate the generalization of our proposed routing model across multiple domains on RouterBench \cite{hu2024routerbench}. It is a large-scale, public dataset designed to evaluate multi-LLM routing systems and contains responses from 11 LLMs on $8$ benchmarks, including MMLU, GSM8K, HellaSwag, ARC Challenge, Winogrande, MBPP and MT-Bench datasets. The dataset contains 1 response per model for each user prompt. However, the same model can answer the same question with multiple different responses. Analysis on RouterBench dataset \cite{hu2024routerbench} suggests that most of the answers that can be answered by an expensive model, can also be answered by smaller models as well. So, a cost-efficient router should learn to discern when a query can be routed to a smaller model. For all proprietary models, we estimate the cost of input and output results based on their API pricing, and Together AI for open-source model. For the datasets MMLU, HellaSwag, GSM8K, ARC Challenge, and Winogrande, responses are evaluated using exact match method, while for MBPP, MT-Bench, and RAG, GPT evaluates responses are further normalized to unit scale.


\paragraph{Baselines.} We compare our proposed predictive router with established baseline routers \cite{hu2024routerbench} including multi-layer perceptron (MLP), support vector machines (SVM) and K nearest neighbors (KNN), as well as other proposed predictive routers. Our gold standard is the oracle router from RouterBench \cite{hu2024routerbench}. We draw parallels with oracle routers using different reward functions to identify the most appropriate formulation (see Section~\ref{sec:results-rewards} and Table~\ref{tab:reward-functions}). Notably, the oracle router based on our exponential rewards achieves the best cost–performance trade-off, routes less queries to the expensive model, and is less sensitive to the user parameter. 
\begin{table}[t]
    \centering
\begin{tabular}{c|c|c|c|c|c|c|c|c}
\toprule
\multirow{2}{*}{LLM Pool} & \multicolumn{2}{|c|}{AIQ $\uparrow$} & \multicolumn{2}{|c|}{$\lambda-\text{sensitivity}_{\text{perf}}$ $\downarrow$} & \multicolumn{2}{|c}{$\lambda-\text{sensitivity}_{\text{cost}}$ $\downarrow$} &\multicolumn{2}{|c}{Max Calls(GPT-4) $\downarrow$}\\
\cline{2-9}
& $R_1$ & $R_2$ & $R_1$ & $R_2$ & $R_1$ & $R_2$ & $R_1$ & $R_2$ \\
\hline
Pool 1 & $0.85616$ & $0.84221$ & 0.0155 & \textbf{0.0035} & 2.66e-05 & \textbf{1.55e-05} & $20.564\%$ & $20.564\%$ \\
Pool 2 & $0.83285$ & $0.83366$ & 0.0213 & \textbf{0.0027} & 8.15e-06 & \textbf{5.24e-06} & $4.165\%$ & $4.165\%$ \\
Pool 3 & $0.87512$ & $0.87362$ & 0.0260 & \textbf{0.0040} & 1.88e-05 & \textbf{1.29e-05} & $10.824\%$ & $10.824\%$ \\
Pool 4 & $0.80434$ & $0.83626$ & 0.0258 & \textbf{0.0019} & 4.74e-05 & \textbf{2.30e-05} & $11.359\%$ & $11.359\%$ \\
\bottomrule
\end{tabular}
\vspace{2 mm}
\caption{Comparison between $R_1$ and $R_2$ oracle routers on the basis of AIQ score, performance and cost sensitivity with $\lambda$, and maximum queries routed to the strongest model. Higher AIQ score and lower queries to the strongest model mean better cost-efficiency, while lower sensitivity indicates robustness of the oracle router to minor variations in $\lambda$, thereby the reward formulation.}
\label{tab:reward-functions}
\end{table}

\paragraph{Evaluation Metrics}
For evaluating cost-efficiency of the routers, we plot a pareto frontier on cost-performance plane \cite{hu2024routerbench} using the average cost vs performance points, obtained by varying user's willingness to pay  ($\lambda$). The area under this convex hull divided by the cost range gives \textbf{Average Improvement over Quality (AIQ)} in Equation~\ref{eq:aiq}, thus aggregating the router's trade-off into a single metric.
Intuitively, higher AIQ score represents better performance is achieved for most of the cost range and vice versa, lower generation cost for most of the performance range. It indicates how well the router trades-off conflicting goals - performance and generation cost, thus serving as our primary metric.

\begin{equation} \label{eq:aiq}
    \text{AIQ}(R) = \frac{1}{b-a} \times \text{Area}_{ \text{cost-perf}}
\end{equation}
Where $[a,b]$ represent the cost range. 
In order to gauge the robustness of reward functions, we perform sensitivity analysis of the reward functions with respect to user's willingness to pay ($\lambda$), that is how abruptly the performance/inference cost varies with $\lambda$. We define \textbf{$\lambda$-sensitivity} with respect to performance as the weighted average of the change in performance over the log scale difference in user parameter ($\lambda$), and cost sensitivity analogously.\footnote{We intentionally keep the performance and cost sensitivity distinct, for fine-grained results and their differing orders of magnitude.} Lower $\lambda$-sensitivity indicates the oracle router remains stable and consistent under small changes in $\lambda$, without severe degradation in performance/cost. 

For instance, if the router's average performance at $\lambda_1$, $\lambda_2$ and $\lambda_3$ user's willingness to pay is $s_1$, $s_2$ and $s_3$ respectively, then $\lambda$-sensitivity with performance is formulated as Equation~\ref{eq:sensitivity}.

\begin{equation} \label{eq:sensitivity}
    \lambda-\text{sensitivity}_{\text{perf}} = \frac{\log{(\lambda_2/\lambda_1)} (s_2 - s_1) + \log{(\lambda_3/\lambda_2)} (s_3 - s_2)}{(\log{(\lambda_3/\lambda_1)})}\\
\end{equation}

Along with AIQ score, we report \textbf{maximum performance} attained over the range of user’s parameter. 
\section{Implementation Details} \label{sec:model-details}

\paragraph{User prompt Embeddings.} User prompts are encoded with DistilBERT into 768-dimensional embeddings, which are subsequently normalized before attention computation.

\paragraph{LLM Representations.} \label{para:llm-representations} 
To capture latent model expertise across domains, we construct fixed-size embeddings for each model. Training queries are first clustered via K-means, and 20\% of prompts are sampled uniformly at random from each cluster as representatives. Given $C$ clusters, a model embedding $\mathcal{I}_\mathbf{m}\in\mathbb{R}^C$ is defined as the mean performance of the model on prompts within each cluster. This training-free approach, inspired by Universal Routing~\cite{jitkrittum2025universal}, uses a large prompt set for clustering to mitigate overfitting risks. Incorporating model embeddings enriches the router’s understanding of model-specific capabilities and improves query–model matching, particularly for complex inputs.



\textbf{Training and Test Details.} 
All predictors are trained with mean squared error (MSE) loss, using the Adam optimizer and CosineAnnealingLR scheduler. The attention-based performance predictor is trained with a learning rate of $1\times 10^{-3}$, batch size of 1024, weight decay of $1\times 10^{-5}$, and 1000 epochs. The attention-based cost predictor uses a learning rate of $1\times 10^{-4}$, batch size of 1024, weight decay of $1\times 10^{-7}$, and maps inputs to an internal dimension of 20. We adopt a 75\%-5\%-20\% train–validation–test split and select hyperparameters based on validation loss.

For LLM representations, the training prompts are clustered into 20 groups (determined by an elbow test), resulting in 20-dimensional model embeddings, while prompt embeddings remain 768-dimensional.

\paragraph{LLM Blender Implementation.}
We implemented the LLM-Blender baseline using the open-source PairRM ranker provided in the LLM-Blender framework~\cite{jiang2023llmblender}. For each prompt in the RouterBench dataset, we obtained responses from all candidate models and constructed all possible response pairs. PairRM was then applied to perform pairwise comparisons, with a “win” assigned to the preferred response in each pair. The model with the highest overall number of wins was selected as the routed model for that prompt. This method requires no additional training and was applied only at test time. Because it relies on outputs from all models, its total cost is computed as the sum of the inference costs of all candidate models per prompt.

\begin{table}
    \centering
\begin{tabular}{c|c|c|c|c|c|c}
\toprule
\multirow{2}{*}{Router} & \multicolumn{2}{|c|}{LLM Pool 1} & \multicolumn{2}{|c|}{LLM Pool 2} & \multicolumn{2}{|c|}{LLM Pool 3} \\
\cline{2-7}
& AIQ $\uparrow$ & $\text{Perf}_{\text{Max}}$ $\uparrow$
& AIQ $\uparrow$ & $\text{Perf}_{\text{Max}}$ $\uparrow$ & AIQ $\uparrow$ & $\text{Perf}_{\text{Max}}$ $\uparrow$\\
\hline
KNN router (k=20) & $\mathit{0.70608}$ & $0.76912$ & $0.49338$ & $0.52573$ & $0.55727$ & $0.65385$ \\
MLP router & $0.67598$ & $0.73781$ & $\mathit{0.66564}$ & $\mathit{0.67551}$ & $\mathit{0.72655}$ & $\mathit{0.76975}$ \\
SVM router (margin=0) & $0.70220$ & $\mathit{0.77233}$ & $0.51452$ & $0.57024$ & $0.49760$ & $0.67767$ \\
Attention router ($R_2$) & $\mathbf{0.72737}$ & $\mathbf{0.78082}$ & $\mathbf{0.66586}$ & $\mathbf{0.67748}$ & $\mathbf{0.74439}$ & $\mathbf{0.78347}$ \\
LLM Blender & - & $0.62314$ & - & $0.58982$ & - & $0.64905$ \\
\bottomrule
\end{tabular}
\vspace{2 mm}
\caption{Comparison of router’s performance and cost-efficiency with traditional routers}
\label{tab:performance-comparison-wth-baselines}
\vspace{-5mm}
\end{table}

\section{Results and Discussion}

\paragraph{Reward functions.} \label{sec:results-rewards} We determine an appropriate reward function in the predictive framework by evaluating oracle routers associated with them. We compare $2$ reward functions, the traditionally used linear trade-off ($R_1$ in Eqn. \eqref{eq:R1}) and novel exponential trade-off ($R_2$ in Eqn. \eqref{eq:R1}), proposing the latter as an appropriate reward function. For a user prompt $q$ with an LLM response $r$, the performance $s(q,r)$ and generation cost $c(q,r)$ are combined in reward functions as:
\begin{equation}
    R_1 = s -\frac{1}{\lambda} c, \; \; \; \; 
    R_2 = s \times \exp\left(-\frac{1}{\lambda} c \right) \label{eq:R1}
\end{equation}

Though both the reward functions have similar AIQ scores (in Table \ref{tab:reward-functions}) while routing at most $20\%$ of user prompts to the expensive LLM, the $\lambda$-sensitivity of $R_2$ oracle router is drastically lesser than that of $R_1$ router, indicating stability of $R_2$ reward function over traditional linear trade-off $R_1$. This could be attributed to the boundedness of the exponential trade-off while linear trade-off is unbounded.

\begin{figure}[t]
    \centering
    \begin{subfigure}{0.32\textwidth}
        \includegraphics[width=\linewidth]{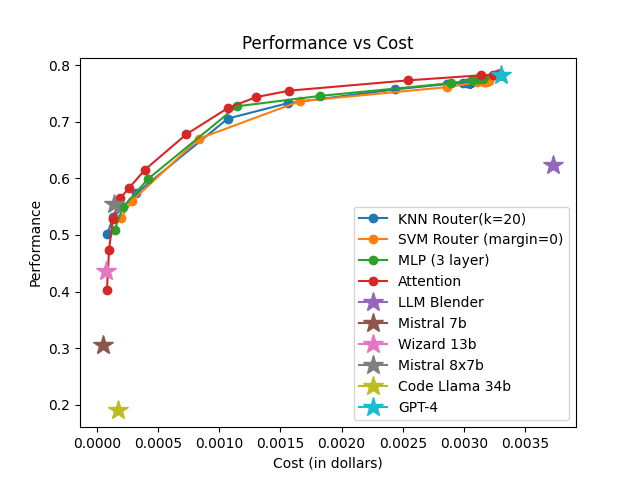}
        \caption{LLM pool 1}
        \label{fig:baselines-pool1}
    \end{subfigure}
    \hfill
    \begin{subfigure}{0.32\textwidth}
        \includegraphics[width=\linewidth]{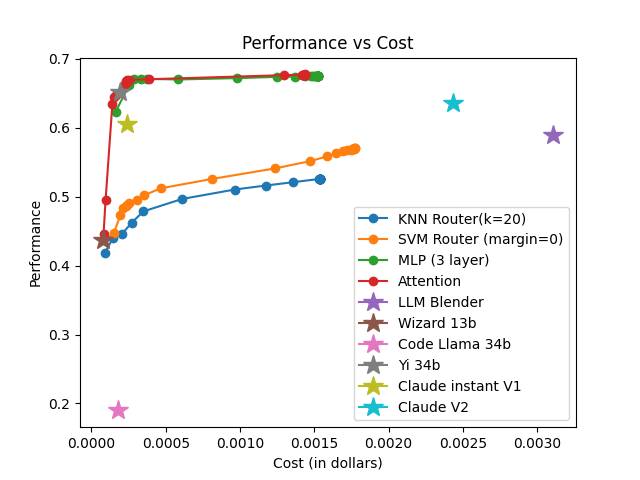}
        \caption{LLM pool 2}
        \label{fig:baselines-pool2}
    \end{subfigure}
    \hfill
    \begin{subfigure}{0.32\textwidth}
        \includegraphics[width=\linewidth]{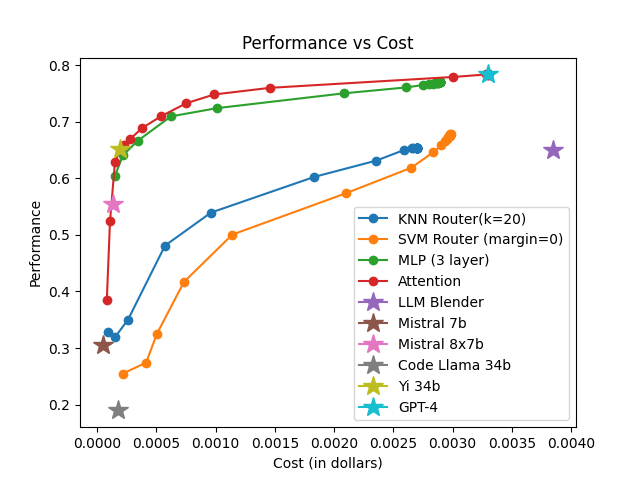}
        \caption{LLM pool 3}
        \label{fig:baselines-pool4}
    \end{subfigure}
    \caption{Comparison of Attention Router with RouterBench Baseline Routers}
    \label{fig:Attnrouter-and-baselines}
\end{figure}

\paragraph{Cost-efficiency of LLM Routers}
Figure~\ref{fig:Attnrouter-and-baselines} and Table~\ref{tab:performance-comparison-wth-baselines} show the performance of our attention-based router against KNN, SVM, and MLP baselines. The key metric is Average Improvement in Quality (AIQ), defined as the area under the cost–quality Pareto frontier. Across all LLM pools, the attention router achieves higher AIQ. In LLM pool 1, it improves AIQ by at least $3\%$ over baselines. In other pools with similarly performing models, it outperforms KNN and SVM by at least $33.58\%$ and $29.41\%$, respectively, along with higher maximum performance ($\text{Perf}_{\text{Max}}$). These results highlight the advantage of our similarity-based routing objective in both performance and cost-efficiency, and demonstrates that attention-based routing provides a robust and generalizable improvement.


\paragraph{Ablation: Different Architectures in predictor-based LLM routing framework}

\begin{figure}[H]
     \centering
     \begin{subfigure}[b]{0.32\textwidth}
         \centering
         \includegraphics[width=\textwidth]{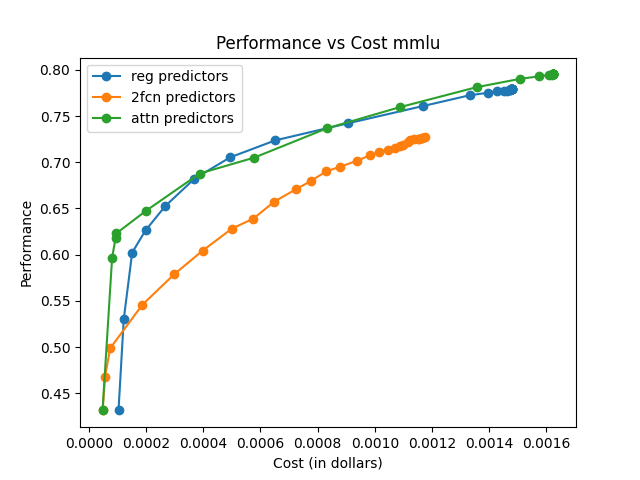}
         \caption{MMLU}
     \end{subfigure}
     \begin{subfigure}[b]{0.32\textwidth}
         \centering
         \includegraphics[width=\textwidth]{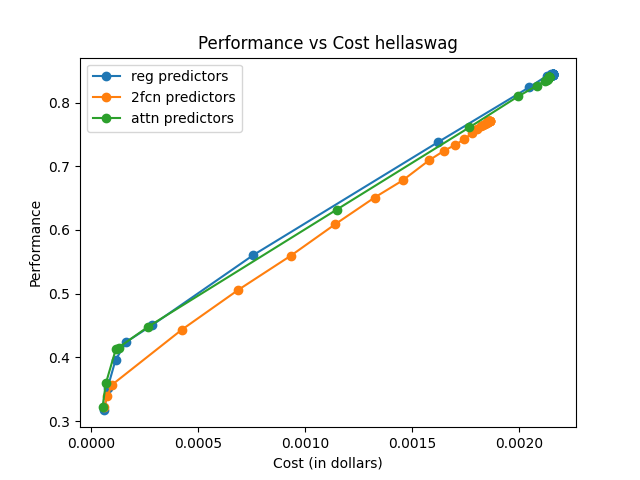}
         \caption{Hellaswag}
     \end{subfigure}
     \begin{subfigure}[b]{0.32\textwidth}
         \centering
         \includegraphics[width=\textwidth]{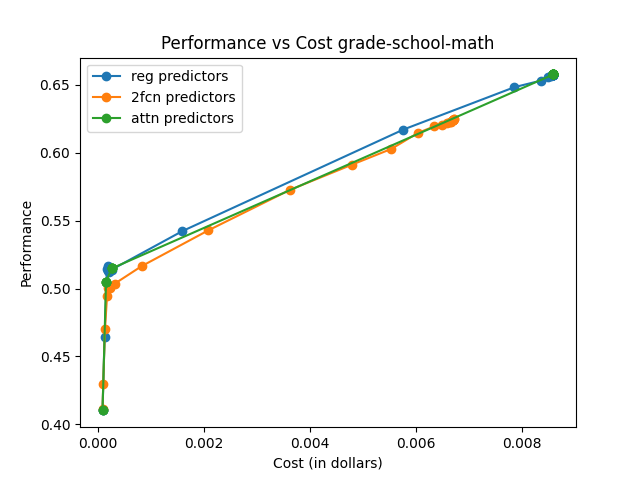}
         \caption{GSM8K}
     \end{subfigure}
     \begin{subfigure}[b]{0.32\textwidth}
         \centering
         \includegraphics[width=\textwidth]{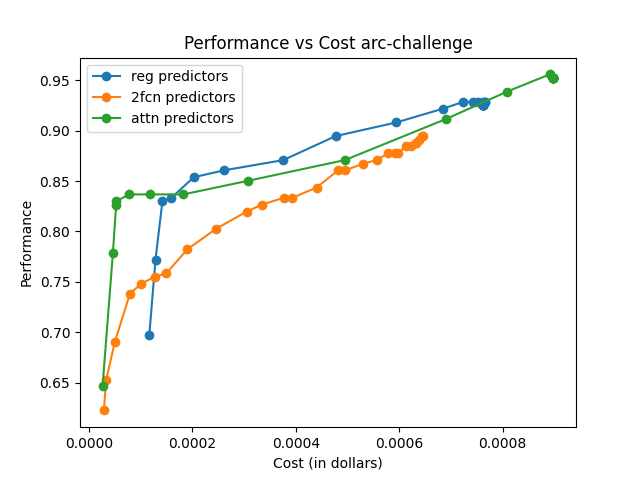}
         \caption{Arc Challenge}
     \end{subfigure}
     \begin{subfigure}[b]{0.32\textwidth}
         \centering
         \includegraphics[width=\textwidth]{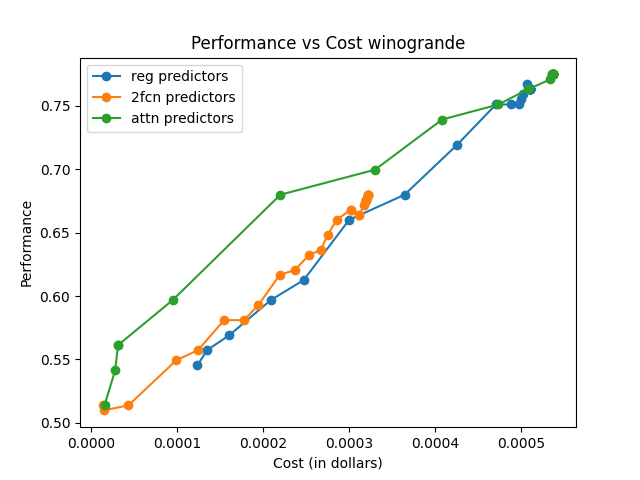}
         \caption{Winogrande}
     \end{subfigure}
     \begin{subfigure}[b]{0.32\textwidth}
         \centering
         \includegraphics[width=\textwidth]{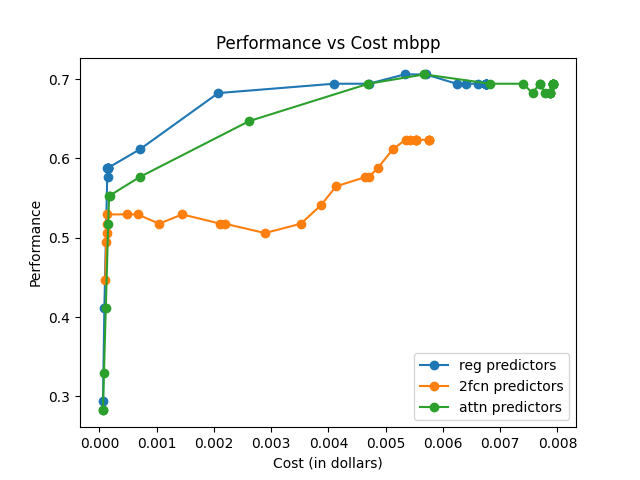}
         \caption{MBPP}
     \end{subfigure}
     \begin{subfigure}[b]{0.32\textwidth}
         \centering
         \includegraphics[width=\textwidth]{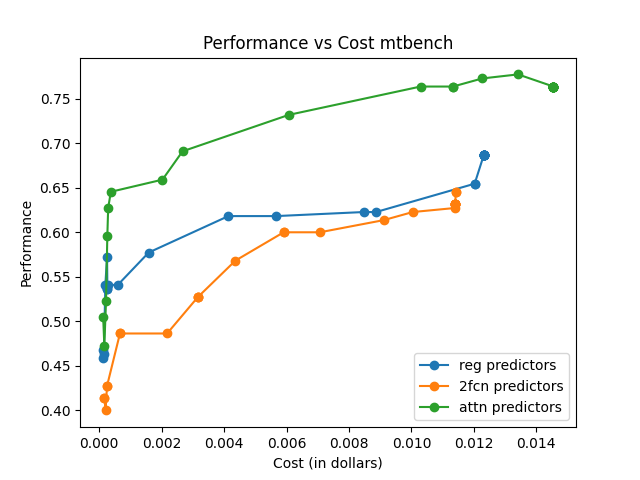}
         \caption{MT Bench}
     \end{subfigure}
     \caption{Dataset-wise results of the predictor-based routers using $R_2$ exponential rewards}
     \label{fig:dataset-wise-r2}
\end{figure}
We ablate predictor architectures (regression, MLP, attention) and domains, finding attention-based router outperforms other predictive routers by up to 6.6\% in AIQ and 2.9\% in maximum performance. Detailed results are presented in Figure~\ref{fig:predictors-comparison-plot} and Tables~\ref{tab:predictor-routing-extra-results-tab1}–\ref{tab:predictor-routing-extra-results-tab4} in Appendix \ref{sec:predictor-routing-extra-results}.

\begin{figure}[H]
     \centering
     \begin{subfigure}[b]{0.32\textwidth}
         \centering
         \includegraphics[width=\textwidth]{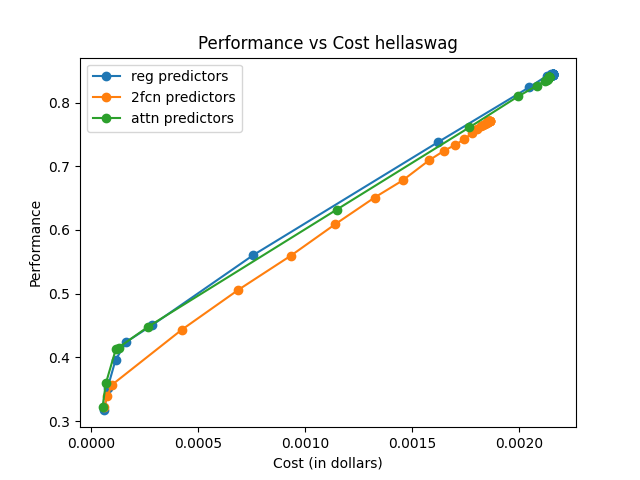}
         \caption{Hellaswag}
     \end{subfigure}
     \begin{subfigure}[b]{0.32\textwidth}
         \centering
         \includegraphics[width=\textwidth]{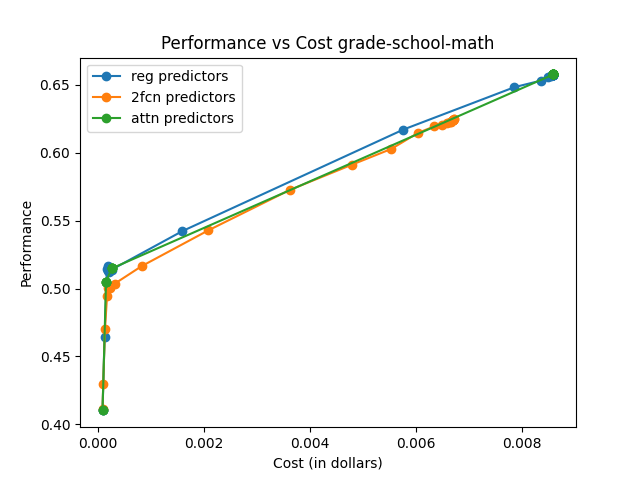}
         \caption{GSM8K}
     \end{subfigure}
     \begin{subfigure}[b]{0.32\textwidth}
         \centering
         \includegraphics[width=\textwidth]{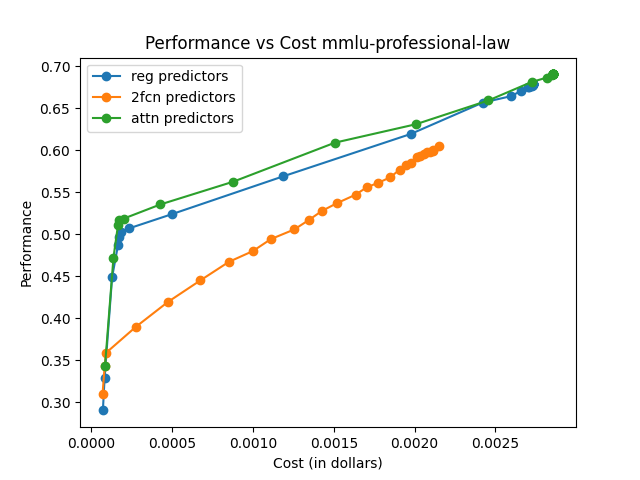}
         \caption{MMLU Professional Law}
     \end{subfigure}
     \begin{subfigure}[b]{0.32\textwidth}
         \centering
         \includegraphics[width=\textwidth]{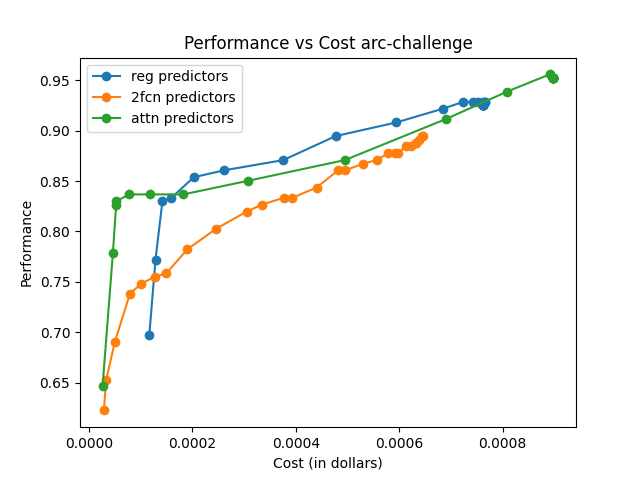}
         \caption{Arc Challenge}
     \end{subfigure}
     \begin{subfigure}[b]{0.32\textwidth}
         \centering
         \includegraphics[width=\textwidth]{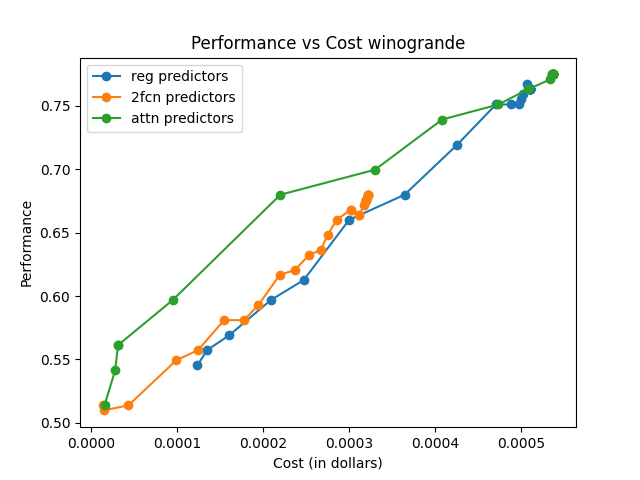}
         \caption{Winogrande}
     \end{subfigure}
     \begin{subfigure}[b]{0.32\textwidth}
         \centering
         \includegraphics[width=\textwidth]{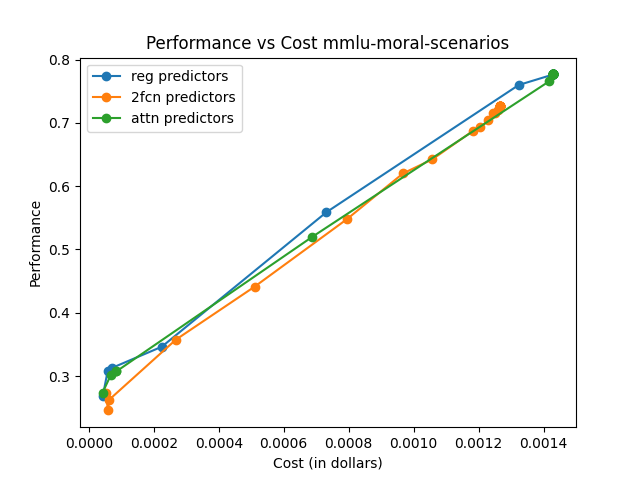}
         \caption{MMLU Moral Scenarios}
     \end{subfigure}
     \begin{subfigure}[b]{0.32\textwidth}
         \centering
         \includegraphics[width=\textwidth]{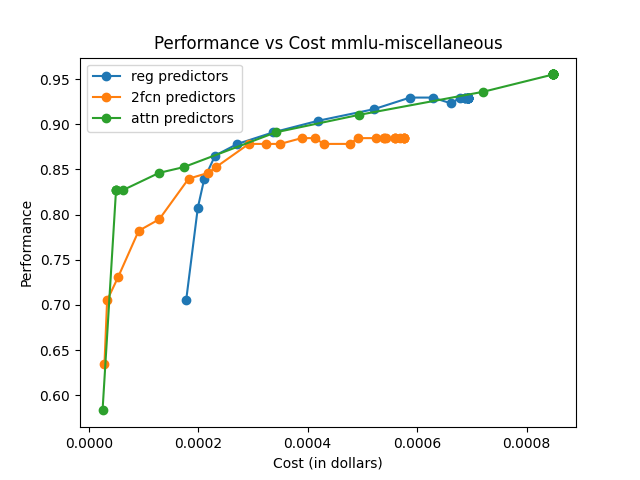}
         \caption{MMLU Miscellaneous}
     \end{subfigure}
     \caption{Domain-wise results of the predictor-based routers using $R_2$ exponential rewards}
     \label{fig:domain-wise-r2}
\end{figure}

Figures~\ref{fig:dataset-wise-r2}--\ref{fig:domain-wise-r2} present results on different datasets and domains, including MMLU, HellaSwag, GSM8K, ARC Challenge, Winogrande, MBPP, and MT-Bench on LLM pool 1 using $R_2$ rewards. (See Figures~\ref{fig:dataset-wise-r1}--\ref{fig:domain-wise-r1} for results using $R_1$ rewards) Across most of the domains, Attention Router consistently matches or exceeds the performance of traditional predictors at lower cost, demonstrating robust generalization, including complex tasks such as ARC Challenge. Particularly, in domains with high diversity (e.g., Winogrande, MT Bench and MMLU Professional Law), our method maintains strong cost--performance trade-offs, validating its adaptability.



\section{Conclusion}
This paper presents a predictor-based LLM routing framework with cross-attention similarity module to optimize cost-performance trade-offs in multi-model environments. By modeling query-model interactions and introducing an exponential reward formulation, our method delivers robust and scalable routing across LLM pools. Empirical results on RouterBench confirm that attention-based predictors outperform traditional baselines in both efficiency and quality, achieving competitive AIQ scores and strong generalization across domains and tasks. These findings highlight the effectiveness of attention-driven routing for large-scale LLM deployment, offering a practical solution for balancing computational cost and response quality in dynamic, real-world settings.
\bibliographystyle{unsrt}  
\bibliography{One_Head_Many_Models_Cross-Attention_Routing_for_Cost-Aware_LLM_Selection}  

\newpage
\appendix
\onecolumn

\section{Proposed Reward Functions' Analysis}
\label{sec:appendix}
\begin{figure}[H]
     \centering
     \begin{subfigure}[b]{0.45\textwidth}
         \centering
         \includegraphics[width=\textwidth]{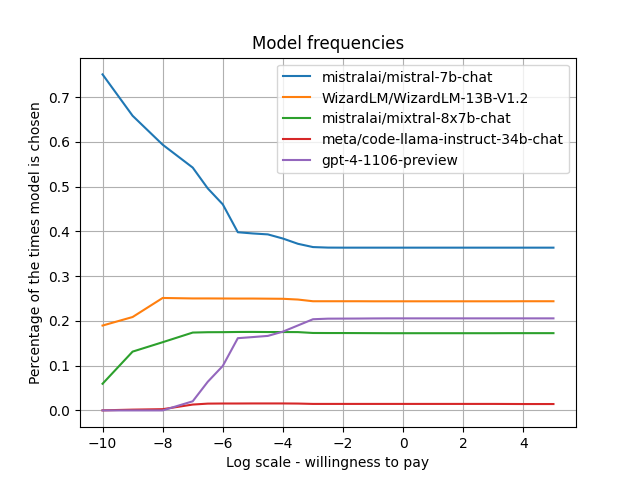}
         \caption{$R1=s(q, r) - \frac{1}{\lambda} c(q, r)$}
         \label{fig:R1-reward-function-analysis}
     \end{subfigure}
     \hfill
     \begin{subfigure}[b]{0.45\textwidth}
         \centering
         \includegraphics[width=\textwidth]{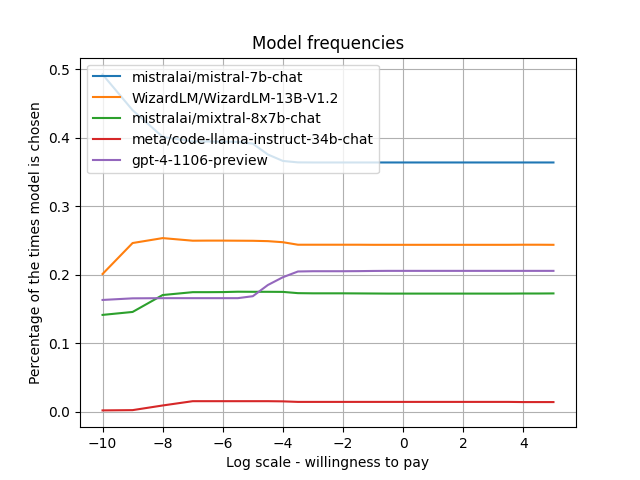}
         \caption{$R2=s(q, r) \exp(-\frac{1}{\lambda} c(q, r))$}
         \label{fig:R21-reward-function-analysis}
     \end{subfigure}
     \caption{Distribution of queries routed to each model in LLM pool 1 by oracle routers with our proposed reward functions}
\end{figure}

The oracle routers we defined in section \ref{sec:method} are ideal routers. As it can be seen in the above plot, these routers not only attain the best performance-cost trade-off, on par with oracle router in RouterBench but also route most of the queries to a lower cost model, while achieving this trade-off. Maximum number of queries routed to GPT-4 with either of the baselines is $~20\%$, thus verifying that employing these reward functions is an appropriate approach for cost-efficient LLM router.







\section{LLM Pools}
We conducted experiments on the following LLM pools: \\
\textbf{LLM pool 1:} Mistral 7B Chat, WizardLM 13B V1.2, Mistral 8x7B Chat, Code Llama Instruct 34B Chat, GPT 4 \\
\textbf{LLM Pool 2:} WizardLM 13B V1.2, Code Llama Instruct 34B Chat, Yi 34B Chat, Claude Instant V1, Claude V2 \\
\textbf{LLM Pool 3:}  Mistral 7B Chat, Mistral 8x7B Chat, Code Llama Instruct 34B Chat, Yi 34B Chat, GPT 4 \\
\textbf{LLM Pool 4:} Llama 2 70B, Claude V1, Claude V2, GPT-4\\

\section{Predictor-based LLM Routing Framework}
\label{sec:predictor-routing-extra-results}
We conducted an ablation study by varying predictors in the predictor-based routing framework. We observe that the router with attention module as both performance and cost predictors yields the best AIQ score and maximum performance.

\subsection{Rewards: $R_1$} 
\begin{table}[H]
\centering
\resizebox{\textwidth}{!}{%
\begin{tabular}{ cc|c|c|c|c|c|c|c|c } 
\hline
\hline
 & & \multicolumn{8}{c}{Cost Predictor} \\
\cline{3-10}
 & & Oracle R1 & Reg & 2-FCN & 3-FCN & Reg-emb & 2-FCN-emb & 3-FCN-emb & Attn \\
 \hline
\multirow{7}{4em}{Quality Predictors} 
& Oracle R1 & $0.85639$ & $0.85442$ & $0.85637$ & $0.85638$ & $0.85771$ & $0.85572$ & $0.85512$ & $0.85830$\\
& Reg & $0.72045$ & $0.71810$ & $0.72090$ & $\mathbf{0.72122}$ & $0.70643$ & $0.72013$ & $0.72007$  & $0.71881$\\ 
& 2-FCN & $0.66528$ & $0.66505$ & $0.66563$ & $0.64657$ & $0.65330$ & $0.66604$ & $\mathbf{0.66637}$ & $0.66628$ \\ 
& 3-FCN & $0.67217$ & $0.67438$ & $0.67415$ & $0.65023$ & $0.66863$ & $0.67370$ & $\mathbf{0.68394}$ & $0.67145$ \\
& Reg-emb & $0.72144$ & $\mathbf{0.72313}$ & $0.72225$ & $0.72100$ & $0.25780$ & $0.72127$ & $0.72099$ & $0.72190$ \\
& 2-FCN-emb & $0.69368$ & $0.69317$ & $0.69382$ & $0.69399$ & $0.67951$ & $\mathbf{0.69501}$ & $0.69417$ & $0.69308$\\
& 3-FCN-emb & $0.68270$ & $0.68211$ & $0.68247$ & $0.68332$ & $0.67491$ & $\mathbf{0.68427}$ & $0.68359$ & $0.68361$\\
& Attn & $0.72540$ & $\mathbf{0.72485}$ & $\mathbf{0.72485}$ & $\mathbf{0.72426}$ & $\mathbf{0.71490}$ & $\mathbf{0.72365}$ & $\mathbf{0.72396}$ & \textcolor{red}{$\mathbf{0.72644}$}  \\
\hline
\hline
\end{tabular}}
\caption{AIQ scores}
\label{tab:predictor-routing-extra-results-tab1}
\end{table}
\begin{figure}[t]
    \centering
    \begin{subfigure}{0.32\textwidth}
        \includegraphics[width=\linewidth]{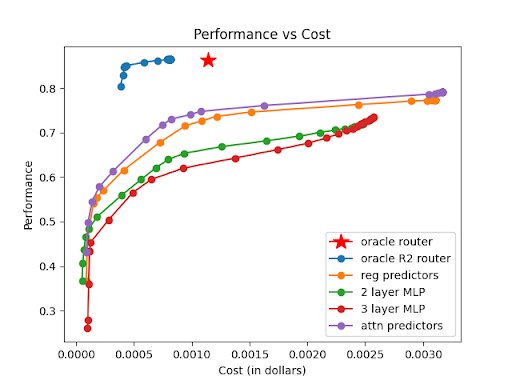}
        \caption{LLM pool 1}
        \label{fig:predictive_routers_pool1}
    \end{subfigure}
    \hfill
    \begin{subfigure}{0.32\textwidth}
        \includegraphics[width=\linewidth]{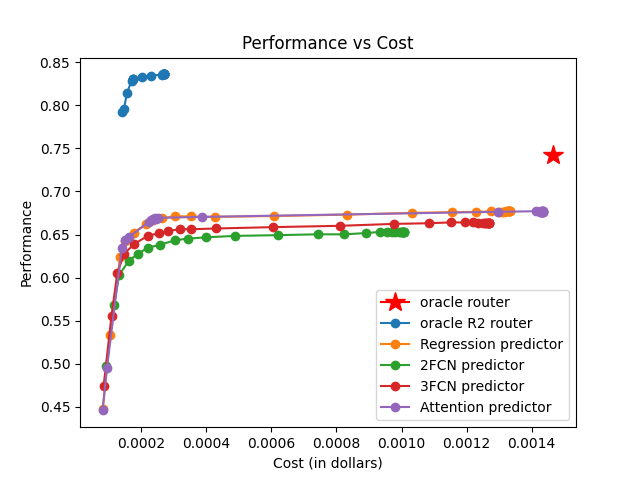}
        \caption{LLM pool 2}
        \label{fig:predictive_routers_pool2}
    \end{subfigure}
    \hfill
    \begin{subfigure}{0.32\textwidth}
        \includegraphics[width=\linewidth]{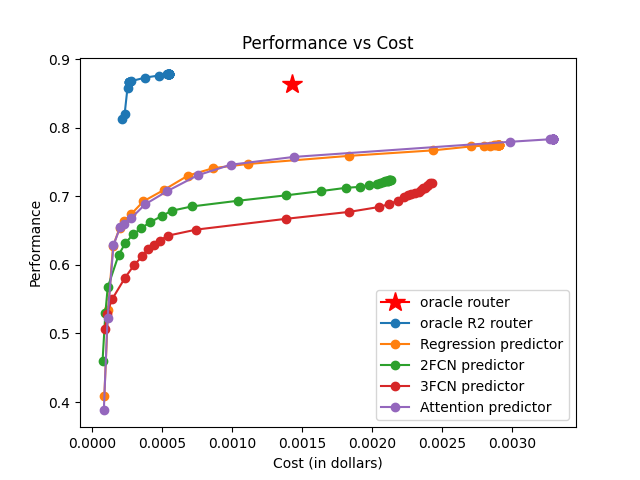}
        \caption{LLM pool 3}
        \label{fig:predictive_routers_pool4}
    \end{subfigure}
    \caption{Cost-efficiency of predictors in predictor-based routing framework}
    \label{fig:predictors-comparison-plot}
\end{figure}

\begin{table}[!h]
\centering
\resizebox{\textwidth}{!}{%
\begin{tabular}{ cc|c|c|c|c|c|c|c|c } 
\hline
\hline
 & & \multicolumn{8}{c}{Cost Predictor} \\
 \cline{3-10}
 & & Oracle R1 & Reg & 2-FCN & 3-FCN & Reg-emb & 2-FCN-emb & 3-FCN-emb & Attn \\
\hline
\multirow{7}{4em}{Quality Predictors} 
& Oracle R1 & $0.86430$ & $0.86430$ & $0.86430$ & $0.86430$ & $0.86430$ & $0.86430$ & $0.86430$ & $0.86430$ \\
& Reg & $0.77338$ & $0.77338$ & $0.77338$ & $0.77338$ & $0.77338$ & $0.77338$ & $0.77338$ & $0.77338$\\ 
& 2-FCN & $0.72036$ & $0.72036$ & $0.72036$ & $0.72039$ & $0.72000$ & $0.72036$ & $0.72036$ & $0.72050$ \\ 
& 3-FCN & $0.73552$ & $0.73562$ & $0.73564$ & $0.73504$ & $0.73581$ & $0.73575$ & $0.76401$ & $0.73601$ \\
& Reg-emb & $0.78337$ & $\mathbf{0.78337}$ & $\mathbf{0.78337}$ & $\mathbf{0.78337}$ & $\mathbf{0.78337}$ & $\mathbf{0.78337}$ & $\mathbf{0.78337}$ & $\mathbf{0.78337}$ \\
& 2-FCN-emb & $0.76811$ & $0.76799$ & $0.76799$ & $0.76799$ & $0.76787$ & $0.76794$ & $0.76812$ & $0.76812$\\
& 3-FCN-emb & $0.76389$ & $0.76402$ & $0.76402$ & $0.76389$ & $0.76420$ & $0.76375$ & $0.76415$ & $0.76389$ \\
& Attn & $0.78082$ & $\mathbf{0.78094}$ &  $\mathbf{0.78082}$ & $\mathbf{0.78091}$ & $\mathbf{0.78082}$ & $\mathbf{0.78082}$ & $\mathbf{0.78082}$ & $\mathbf{0.78082}$\\
\hline
\hline
\end{tabular}}
\caption{Maximum performance achieved}
\label{tab:predictor-routing-extra-results-tab2}
\end{table}

\subsection{Rewards: $R_2$}

\begin{table}[H]
\centering
\resizebox{\textwidth}{!}{%
\begin{tabular}{ c c|c|c|c|c|c|c|c|c } 
\hline
\hline
 & & \multicolumn{8}{c}{Cost Predictor} \\
 \cline{3-10}
 & & Oracle R2 & Reg & 2-FCN & 3-FCN & Reg-emb & 2-FCN-emb & 3-FCN-enb & Attn \\
\hline
\multirow{7}{4em}{Quality Predictors} 
& Oracle R2 & $0.84275$ & $0.84518$ & $0.84361$ & $0.84318$ & $0.85961$ & $0.85378$ & $0.85307$ & $0.85564$ \\
& Reg & $0.72122$ & $0.71833$ & $0.72129$ & $0.72146$ & $0.70525$ & $0.72133$ & $0.72127$ & $\mathbf{0.71949}$ \\ 
& 2-FCN & $0.66427$ & $0.66404$ & $0.66444$ & $0.66348$ & $0.65120$ & $0.66484$ & $0.66517$ & $\mathbf{0.66545}$ \\ 
& 3-FCN & $0.66970$ & $0.67162$ & $0.67050$ & $0.67037$ & $0.66983$ & $\mathbf{0.67180}$ & $0.66887$ & $0.66857$ \\
& Reg-emb & $0.72229$ & $0.72244$ & $\mathbf{0.72274}$ & $0.72249$ & $0.70136$ & $0.72258$ & $0.72249$ & $0.71932$ \\
& 2-FCN-emb & $0.69053$ & $0.68990$ & $0.69038$ & $0.69093$ & $0.67823$ & $\mathbf{0.69197}$ & $0.69011$ & $0.68990$ \\
& 3-FCN-emb & $0.68282$ & $0.68314$ & $0.68347$ & $0.68429$ & $0.67701$ & $\mathbf{0.68465}$ & $0.68447$ & $0.68388$ \\
& Attn-eval & $0.72433$ & $\mathbf{0.72340}$ & $\mathbf{0.72430}$ & $\mathbf{0.72476}$ & $\mathbf{0.71189}$ & $\mathbf{0.72307}$ & $\mathbf{0.72328}$ & \textcolor{red}{$\mathbf{0.72737}$} \\
\hline
\hline
\end{tabular}}
\caption{AIQ scores}
\label{tab:predictor-routing-extra-results-tab3}
\end{table}

\begin{table}[!h]
\centering
\resizebox{\textwidth}{!}{%
\begin{tabular}{ c c|c|c|c|c|c|c|c|c } 
\hline
\hline
 & & \multicolumn{8}{c}{Cost Predictor} \\
 \cline{3-10}
 & & Oracle R2 & Reg & 2-FCN & 3-FCN & Reg-emb & 2-FCN-emb & 3-FCN-enb & Attn \\
\hline
\multirow{7}{4em}{Quality Predictors} 
& Oracle R2 & $0.86430$ & $0.86430$ & $0.86430$ & $0.86430$ & $0.86430$ & $0.86430$ & $0.86430$ & $0.86430$\\
& Reg & $0.77338$ & $0.77338$ & $0.77338$ & $0.77338$ & $0.77338$ & $0.77338$ & $0.77338$ & $0.77338$ \\ 
& 2-FCN & $0.72036$ & $0.72036$ & $0.72036$ & $0.72022$ & $0.72000$ & $0.72036$ & $0.72036$ & $0.72050$ \\ 
& 3-FCN & $0.73526$ & $0.73565$ & $0.73526$ & $0.73578$ & $0.73594$ & $0.73604$ & $0.73530$ & $0.73604$\\ 
& Reg-emb & $0.78337$ & $\mathbf{0.78337}$ & $\mathbf{0.78337}$ & $\mathbf{0.78337}$ & $\mathbf{0.78337}$ & $\mathbf{0.78337}$ & $\mathbf{0.78337}$ & $\mathbf{0.78337}$ \\
& 2-FCN-emb & $0.76824$ & $0.76812$ & $0.76812$ & $0.76812$ & $0.76800$ & $0.76807$ & $0.76825$ & $0.76825$\\
& 3-FCN-emb & $0.76389$ & $0.76402$ & $0.76401$ & $0.76402$ & $0.76425$ & $0.76376$ & $0.76415$  & $0.76402$\\
& Attn-eval & $0.78082$ & $\mathbf{0.78082}$ & $\mathbf{0.78082}$ & $\mathbf{0.78082}$ & $\mathbf{0.78082}$ & $\mathbf{0.78082}$ & $\mathbf{0.78082}$ & $\mathbf{0.78082}$ \\
\hline
\hline
\end{tabular}}
\caption{Maximum performance achieved}
\label{tab:predictor-routing-extra-results-tab4}
\end{table}

\section{Dataset and domain wise results}
\label{sec:domain-dataset-wise}

\begin{figure}[H]
     \centering
     \begin{subfigure}[b]{0.32\textwidth}
         \centering
         \includegraphics[width=\textwidth]{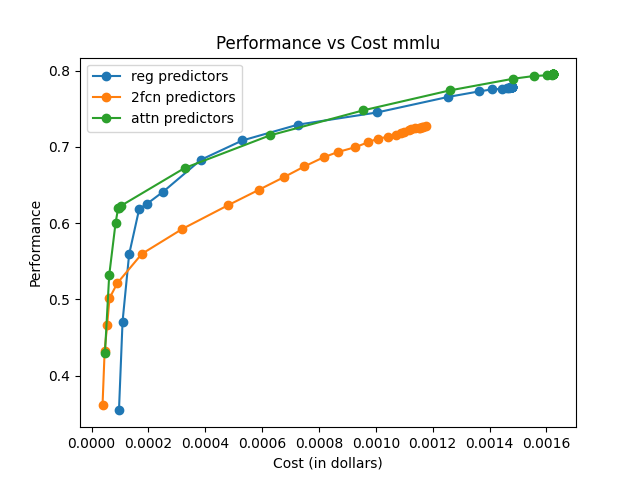}
         \caption{MMLU}
     \end{subfigure}
     \begin{subfigure}[b]{0.32\textwidth}
         \centering
         \includegraphics[width=\textwidth]{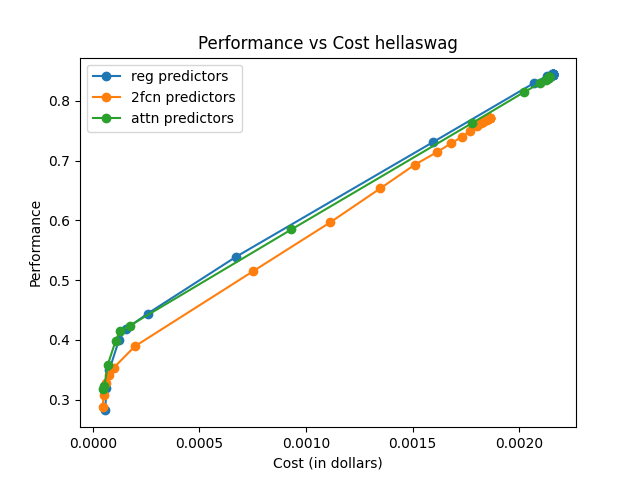}
         \caption{Hellaswag}
     \end{subfigure}
     \begin{subfigure}[b]{0.32\textwidth}
         \centering
         \includegraphics[width=\textwidth]{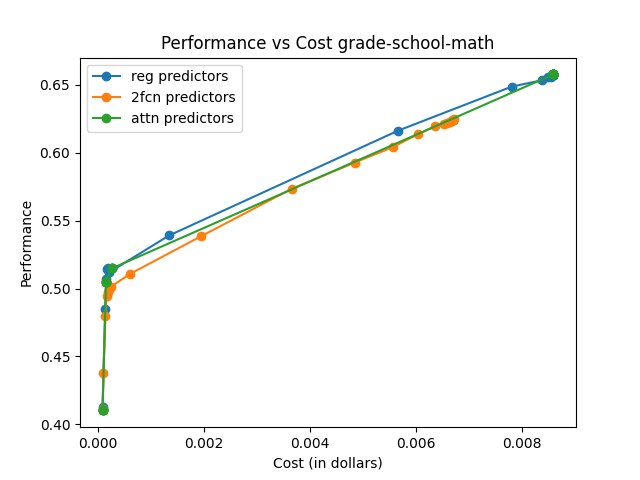}
         \caption{GSM8K}
     \end{subfigure}
     \begin{subfigure}[b]{0.32\textwidth}
         \centering
         \includegraphics[width=\textwidth]{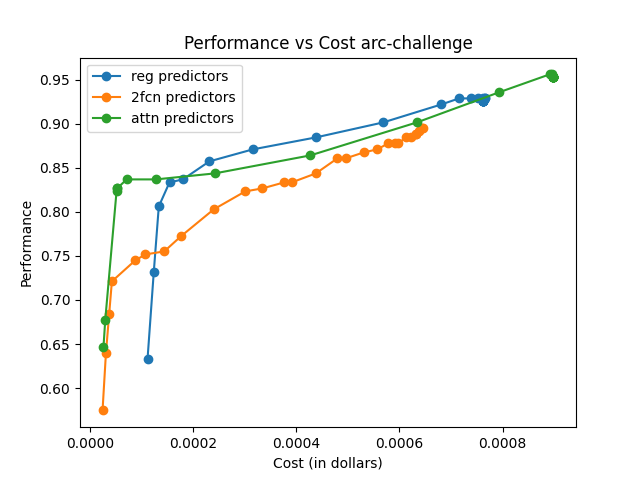}
         \caption{Arc Challenge}
     \end{subfigure}
     \begin{subfigure}[b]{0.32\textwidth}
         \centering
         \includegraphics[width=\textwidth]{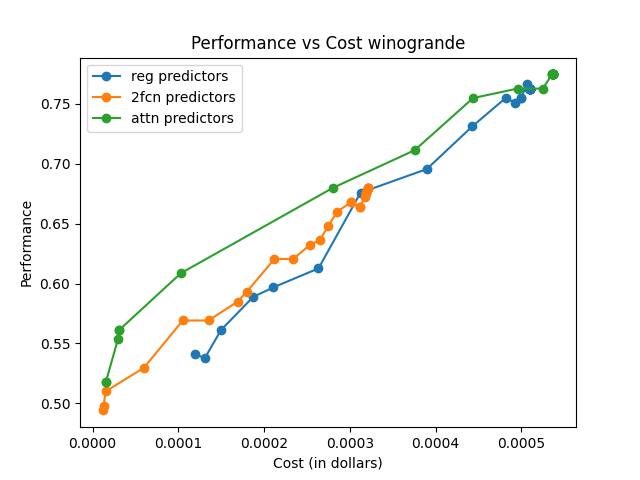}
         \caption{Winograde}
     \end{subfigure}
     \begin{subfigure}[b]{0.32\textwidth}
         \centering
         \includegraphics[width=\textwidth]{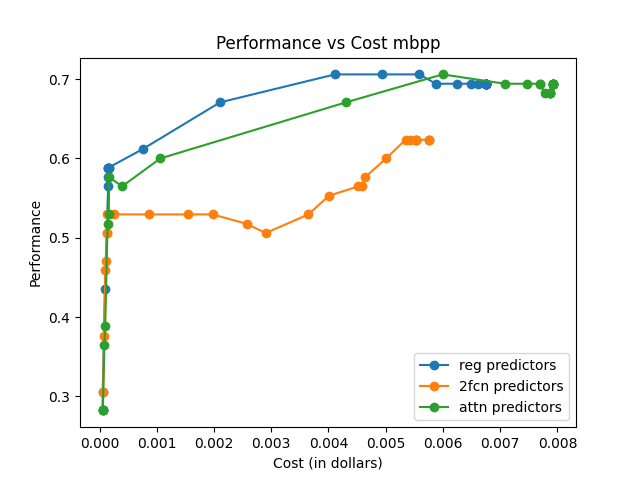}
         \caption{MBPP}
     \end{subfigure}
     \begin{subfigure}[b]{0.32\textwidth}
         \centering
         \includegraphics[width=\textwidth]{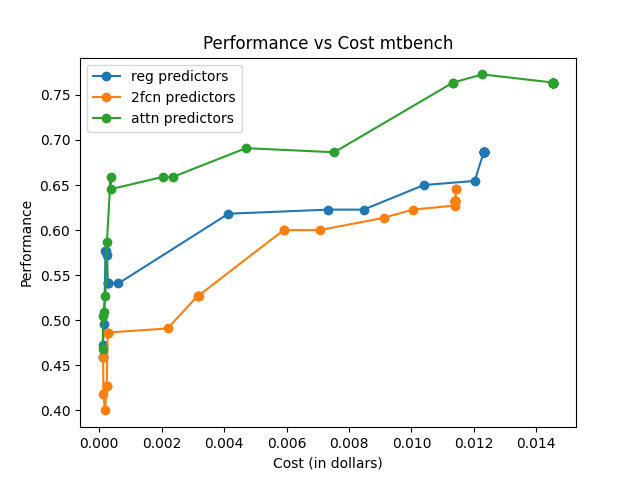}
         \caption{MT Bench}
     \end{subfigure}
     \caption{Dataset-wise results of the predictor-based routers using $R_1=s(q, r) - \frac{1}{\lambda} c(q, r)$ rewards}
     \label{fig:dataset-wise-r1}
\end{figure}

\begin{figure}[H]
     \centering
     \begin{subfigure}[b]{0.32\textwidth}
         \centering
         \includegraphics[width=\textwidth]{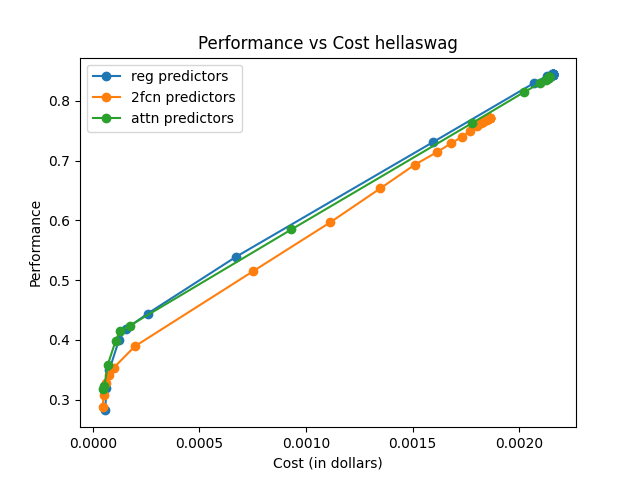}
         \caption{Hellaswag}
     \end{subfigure}
     \begin{subfigure}[b]{0.32\textwidth}
         \centering
         \includegraphics[width=\textwidth]{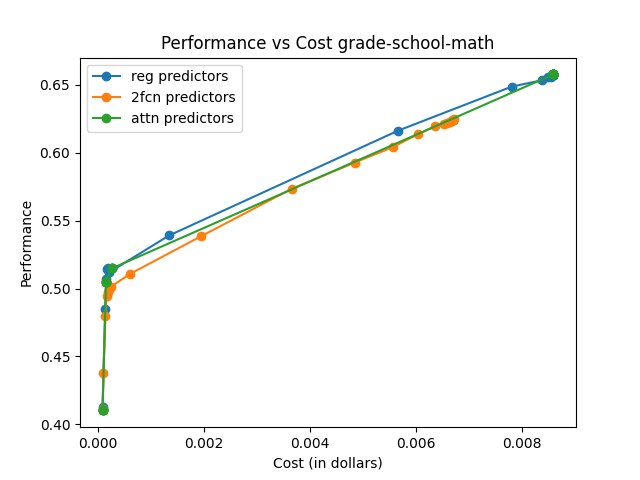}
         \caption{GSM8K}
     \end{subfigure}
     \begin{subfigure}[b]{0.32\textwidth}
         \centering
         \includegraphics[width=\textwidth]{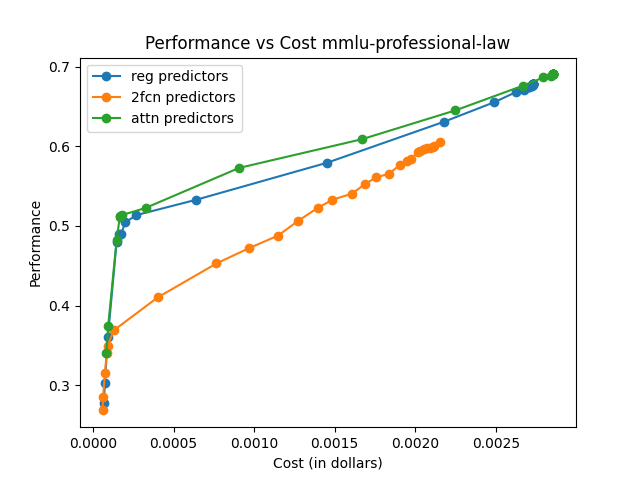}
         \caption{MMLU Professional Law}
     \end{subfigure}
     \begin{subfigure}[b]{0.32\textwidth}
         \centering
         \includegraphics[width=\textwidth]{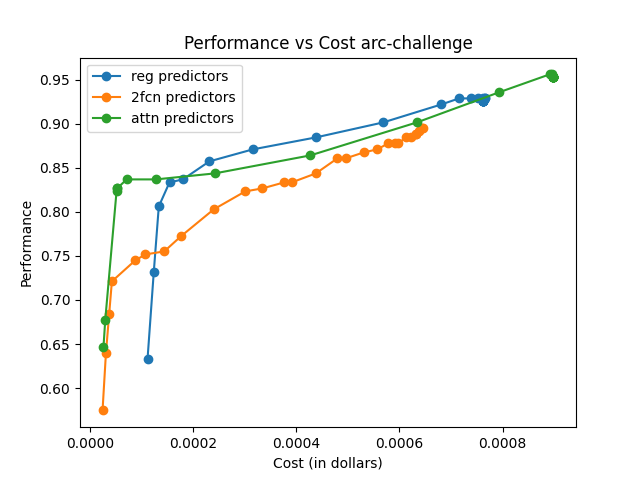}
         \caption{Arc Challenge}
     \end{subfigure}
     \begin{subfigure}[b]{0.32\textwidth}
         \centering
         \includegraphics[width=\textwidth]{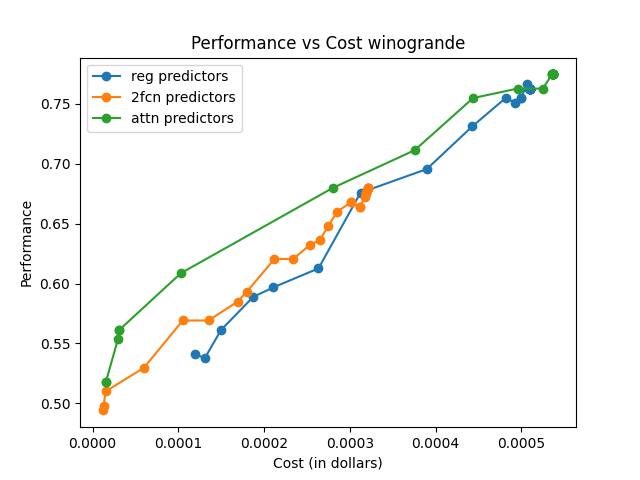}
         \caption{Winograde}
     \end{subfigure}
     \begin{subfigure}[b]{0.32\textwidth}
         \centering
         \includegraphics[width=\textwidth]{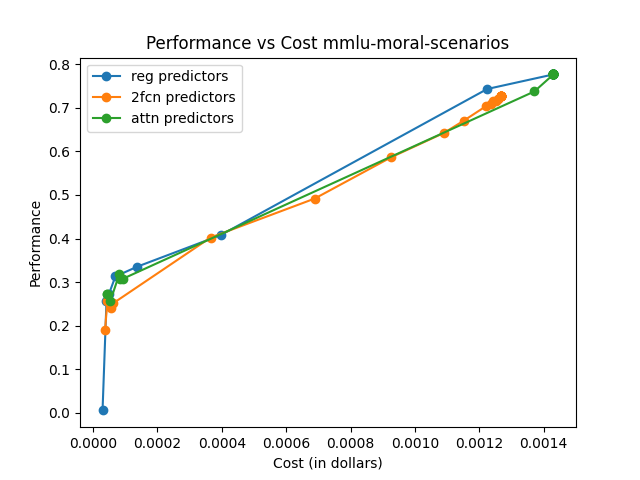}
         \caption{MMLU Moral Scenarios}
     \end{subfigure}
     \begin{subfigure}[b]{0.32\textwidth}
         \centering
         \includegraphics[width=\textwidth]{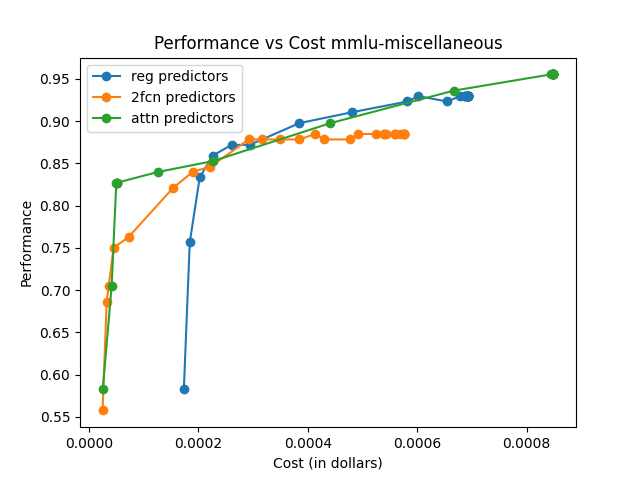}
         \caption{MMLU Miscellaneous}
     \end{subfigure}
     \caption{Domain-wise results of the predictor-based routers using $R_1=s(q, r) - \frac{1}{\lambda} c(q, r)$ rewards}
     \label{fig:domain-wise-r1}
\end{figure}

This section presents cost-performance trade-offs across different datasets and domains with $R_1$ reward function in Figures~\ref{fig:dataset-wise-r1}
--\ref{fig:domain-wise-r1}.

\subsection{Dataset-wise results}
\label{sec:dataset-wise}

\clearpage

\subsection{Domain-wise results}
\label{sec:domain-wise}

\section{Compute Resources} \label{sec:resources}
We implemented our proposed method in PyTorch and conducted experiments on either a single NVIDIA A40 or a single NVIDIA A100 Tensor Core GPU. Our models consume at most 1GB memory. While the training period of predictors for the framework is upto 30 minutes, inference time for predictors is around 5-10 minutes, varying with batch size, trainset size and the architecture.

\section{Limitations} \label{sec:limitations}

Our work does not include direct experimental comparisons with recent multi-LLM routing methods on RouterBench, such as Universal Model Routing. Future work should benchmark our approach against these contemporary baselines for a more robust evaluation. Moreover, our experiments are conducted on a static dataset with fixed LLM pool and tasks from RouterBench. There is a degree of uncertainty with the same LLM responses for the same query, so there is a scope of making it dynamic, and modeling performance and inference cost with a degree of uncertainty. Finally, the method relies on LLM representations, and results vary based on the quality and quantity of the data.

\end{document}